\documentclass[runningheads]{llncs}
\usepackage[T1]{fontenc}
\usepackage{amsmath}
\usepackage{multirow}
\usepackage{xcolor}
\usepackage{graphicx}
\begin{document}
\large This work has been submitted to a conference or journal for possible publication. Copyright may be transferred without notice, after which this version may no longer be accessible.

\newpage
\title{Scene 3-D Reconstruction System in Scattering Medium}

%
%
\author{Zhuoyifan Zhang\inst{1}\and
Lu Zhang\inst{2,4}\and
Liang Wang\inst{3,4} \and
Haoming Wu\inst{2,4,*} 
}
\authorrunning{Zhang Z et al.}
%
\institute{
School of Artificial Intelligence, Beijing University of Posts and Telecommunications, Beijing, China\and
School of Information and Communication Engineering, Hainan University, Haikou, China\and
School of Computer Science and Technology, Hainan University, Haikou, China \and
RobAI-Lab, Hainan University 
\\
\email{\{comzhangzhuoyifan\}@bupt.edu.cn,\{LazLu666,wl3141624820,haomingwu718\}@gmail.com }
}
\maketitle              
\begin{abstract}
The research on neural radiance fields for new view synthesis has experienced explosive growth with the development of new models and extensions. The NERF algorithm, suitable for underwater scenes or scattering media, is also evolving. Existing underwater 3D reconstruction systems still face challenges such as extensive training time and low rendering efficiency. This paper proposes an improved underwater 3D reconstruction system to address these issues and achieve rapid, high-quality 3D reconstruction.To begin with, we enhance underwater videos captured by a monocular camera to correct the poor image quality caused by the physical properties of the water medium while ensuring consistency in enhancement across adjacent frames. Subsequently, we perform keyframe selection on the video frames to optimize resource utilization and eliminate the impact of dynamic objects on the reconstruction results. The selected keyframes, after pose estimation using COLMAP, undergo a three-dimensional reconstruction improvement process using neural radiance fields based on multi-resolution hash coding for model construction and rendering.

\end{abstract}
\section{Introduction}
The Neural Radiance Fields (NeRF) \cite{mildenhall2021nerf} proposed by Mildenhall et al. has attracted wide attention and influence in the field of computer graphics. Based on the ideas of deep learning and neural networks, it uses neural network models to model and represent object surfaces in three-dimensional scenes. Compared to traditional graphics rendering methods, NeRF has a higher level of detail accuracy and precision.\\
The application of NeRF to the 3D reconstruction of underwater scenes is of great significance to the development and management of underwater resources, Marine scientific research and protection, and the development of Marine tourism. Through underwater 3D reconstruction techniques, researchers gain deep insights into the topography, ecosystems, cultural heritage, and infrastructure of underwater worlds, providing robust support for scientific research, environmental conservation, and human activities. Thus, this field holds both scientific value and extensive practical applications.But because NeRF uses neural networks to model the characteristics of a scenario, both its training and reasoning processes require significant computational resources and time. Training a high-quality NeRF model can require a large number of training samples and hours to days of training time. At the same time, the physical environment targeted by NeRF is a clean air medium. For a medium that absorbs or flashes light, such as water, the volume rendering equation not only has a volume meaning for the object, but also the external environment will affect the rendering.\\
In contrast to clear air conditions, when the medium involves absorption or scattering (e.g., haze, fog, smoke, and all aquatic habitats), the volume rendering equation takes on a true volumetric meaning, as the entire volume, not just objects, contributes to the image intensity. Since the NeRF model estimates color and density at every point in the scene, it lends itself to perfect general volume rendering when an appropriate rendering model is used.The choice of water as a scattering medium is due to its prominent light scattering characteristics in image acquisition. The light scattering characteristics in water manifest as the interaction of light rays with water molecules and suspended particles, causing the light to scatter in different directions. The way light propagates in water, including scattering, absorption, and refraction, makes it an ideal model for studying and understanding the behavior of light in scattering media. Due to the relative ease of conducting image capture underwater, water is commonly used as a typical scattering medium for experimental research.\\
Our proposed underwater scene reconstruction system uses an improved 3D reconstruction method of neural radiation field optimized by multi-resolution hash coding \cite{mueller2022instant} to achieve model construction and rendering. This coding method based on hash search only needs a small scale neural network to achieve the effect of a fully connected network without loss of accuracy. Based on a multi-level voxel search structure, the weight search of data is realized, so that the weight optimization and data calculation can be controlled step by step in different levels of the corresponding sub-regions. In this way, for weight optimization, too many ineffective calculation processes can be avoided. In image preprocessing, water's absorption of light is different in different spectral regions and has obvious selectivity. The selective absorption of light by water makes the color of the underwater object change with the increase of its depth. At the same time, when the light propagates in water, it is affected by the medium particles and deviates from the original direction of linear propagation, which is called the scattering of light by water. This phenomenon will reduce the contrast of the image and make the imaging system unable to receive useful information. The image enhancement method we used has good performance on SSIM, UCIQE and UIQM indicators, which can effectively restore the original color of the image and improve the image quality.\\
The main contribution of this paper is to establish a unified 3D reconstruction system for underwater scenes. After image enhancement of underwater video captured by monocular camera, key frames are screened for estimation of pose under COLMAP\cite{schoenberger2016sfm}\cite{schoenberger2016mvs}. Finally, an improved 3D reconstruction method of neural radiation field based on multi-resolution hash coding optimization is used to construct and render the model. The experiment shows that our underwater scene reconstruction method is better than other methods, greatly improving the reconstruction efficiency and ensuring the high-quality reconstruction effect.Compared to existing three-dimensional reconstruction systems for scattering media like Seathru-NERF, our proposed system offers advantages such as shorter required time, superior image enhancement effects, and higher reconstruction quality.\\

\begin{figure}[htbp]
      \centering
      \includegraphics[scale=0.25]{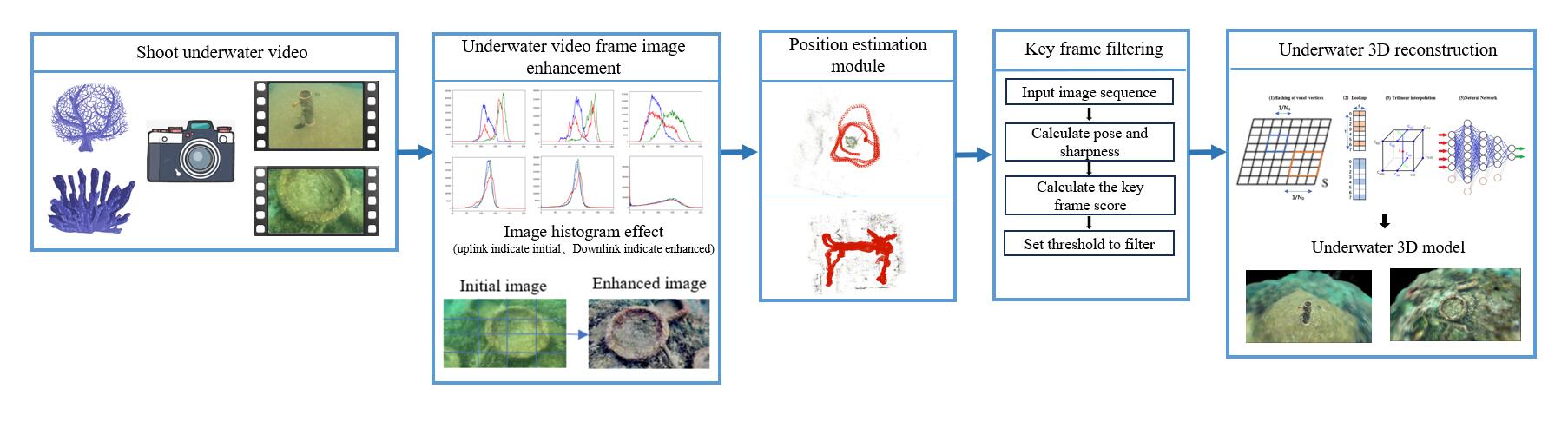}
      \caption{Our method has five parts, image enhancement is the preprocessing part, mainly to enhance the clarity of the underwater image, to facilitate the subsequent 3D reconstruction, the next stage, the position estimation module is to obtain the image position information as an input to the 3D reconstruction, after the next stage of keyframe filtering to conserve the arithmetic resources, and finally 3D reconstruction, which can be rendered from an arbitrary point of view of this 3D model}
\end{figure}
\section{Related Work}
\subsection{Underwater Image Enhancement}
In underwater optical imaging, the inherent properties of water, such as significant light absorption and scattering, result in the exponential attenuation of light propagation underwater. Conventional imaging systems used in underwater settings often suffer from high noise levels, pronounced color aberrations, and distortions, leading to poor image quality. In previous work, numerous underwater image enhancement methods have been proposed to address these challenges.\\
Treibitz et al\cite{6241430}. combined information from different illumination frames to achieve optimal contrast for each region in the output image. Xueyang Fu et al\cite{7025927}. introduced a variational framework for retinex-based image enhancement, effectively addressing color, exposure, and blurriness issues in underwater imaging. Muhammad Suzuri Hitam et al\cite{6522017}. extended the Contrast Limited Adaptive Histogram Equalization (CLAHE) method to underwater image enhancement, utilizing a mixture of CLAHE on both the RGB and HSV color models and combining the results using the Euclidean norm, significantly improving the visual quality of underwater images.\\
Given the effectiveness of the physical models used in underwater image processing and the inherent characteristics of underwater images, we propose a method that combines the CLAHE algorithm with the Retinex enhancement technique for underwater image enhancement. CLAHE processes images in blocks, building upon the adaptive histogram equalization algorithm while introducing a threshold to limit contrast, mitigating the problem of noise amplification. Linear or bilinear interpolation is used to optimize transitions between blocks, resulting in a more harmonious appearance. Multiscale processing is applied to enhance brightness and color representation.\\
The Retinex algorithm, known for its effectiveness in enhancing brightness and contrast in underwater images, decomposes the image into local and global components for enhancement. The local component refers to an adaptive neighborhood for each pixel in the image, while the global component encompasses the entire image. The Retinex algorithm decomposes the value of each pixel into two parts: reflectance and illumination. It then enhances the reflectance values of each pixel to improve image contrast and brightness. Finally, the enhanced reflectance values are multiplied by the illumination values to obtain the final image.
\subsection{Keyframe Filtering}
The selection of keyframes plays a pivotal role in various computer vision and robotics applications, particularly in the context of visual SLAM (Simultaneous Localization and Mapping) and 3-D reconstruction. This section provides an overview of related works in keyframe selection, highlighting the different approaches and strategies employed in the field.\\
Klein and Murray \cite{klein2007parallel} and Mur-Artal et al. \cite{mur2015orb} rely on camera pose information for keyframe selection, often considering motion and loop closure. 
 Konolige and Agrawal  \cite{konolige2008frameslam} introduced mutual information-based keyframe selection.These maximize information gain, using metrics like mutual information.
 Revaud et al. \cite{revaud2015epicflow} proposed quality-driven selection, considering sharpness and overall image quality.
Combining pose and quality criteria, Dubé et al. \cite{dube2017segmatch} aim for accuracy and image quality balance, considering visual saliency and pose for keyframe selection.
Our proposed keyframe selection module extends and innovates upon these existing works by simultaneously considering both pose information and image quality, offering a unique perspective on keyframe filtering in the context of underwater scene reconstruction. In the following sections, we will delve into the details of our approach and present experimental results showcasing its effectiveness.

\subsection{Underwater Reconstruction}
NeRF (Neural Radiance Fields), as a very hot emerging computer vision technology, aims to generate realistic 3D scene reconstruction and rendering.In 2020, Mildenhall et al\cite{mildenhall2021nerf}. first proposed the NeRF technology, which has attracted a lot of attention in academia and industry. Its core idea is to use a neural network to represent the radiation field (radiance field) of each point in a 3D scene, and train the network to estimate the color and depth values of each point. In recent years, a large number of NeRF-related improvements have been proposed, mainly focusing on improving the training efficiency, rendering quality, and expanding the application areas, etc. NeRF$\mathbf{++}$\cite{zhang2020nerf++} by Zhang K et al. further improves the rendering quality and efficiency of NeRF by introducing techniques such as regularization and local feature extraction, etc.Yan Z et al. proposed D-NeRF\cite{pumarola2021d}, which extends the application of NeRF from static scenes to dynamic scenes, and realizes the modeling and rendering of dynamic objects. Based on the consideration of reconstruction efficiency and quality, we use instant-ngp (Instant Neural Graphics Primitives)\cite{mueller2022instant} to realize fast and high-quality 3D reconstruction. instant-ngp proposes a coding method that allows the use of a small-scale network to implement NeRF without loss of accuracy. The network is augmented by a multiresolution hash table of feature vectors, performing optimizations based on stochastic gradient descent. The multi-resolution structure facilitates GPU parallelism and is able to reduce computation by eliminating hash conflicts. The implementation improves NeRF's time overhead in hours to seconds.\\
For 3D reconstruction of scenes with specular effects and reflections, there is also recent challenging work that focuses on optimizing the stability of NeRF as well as expanding the scenarios in which it can be used.NeRF-W\cite{martinbrualla2020nerfw} is able to learn and render reconstructed images containing transient objects, but also guarantees separation from the static network without introducing artifacts to the representation of the static scene. A secondary voxel radiation field incorporating a data-dependent uncertainty field is utilized to reduce the impact of transient objects on the static scene representation.NeRF-ReN\cite{guo2022nerfren} focuses on the 3D reconstruction of reflective scenes, and proposes the use of separate transmission and reflection neural radiation fields in complex reflective scenes. By dividing the scene into transmission and reflection components, a new parametric definition is proposed that can handle reflection as well as specular scenes well.\\
For the 3D reconstruction of underwater scenes, laser scanning, structured light projection, and underwater sonar are currently used. Our proposed method for underwater 3D reconstruction mainly works by preprocessing the optical images captured by a monocular camera, and then the enhanced images are used for subsequent reconstruction work.Seathru-NeRF, proposed by Daniel Levy et al\cite{levy2023seathru}. in 2023, develops a new rendering model for nerf in scattering media based on the SeaThru image imaging model. The idea of Seathru-NeRF's underwater 3D reconstruction method has similarities with our method. In the subsequent 3D reconstruction experimental part, we mainly compare with Seathru-NeRF in terms of rendering quality, training effect, etc., to demonstrate the superiority and efficiency of our method.
\section{A Rapid Underwater Scene Reconstruction System}

\subsection{Underwater Image Enhancement}
Due to issues such as refraction and scattering of light in water, the collected underwater images often suffer from blurriness and a bluish or purplish color cast. To address this problem, our algorithm aims to enhance the original underwater images, improving clarity and color restoration. Conventional enhancement methods often result in significant variations in enhancement effects due to the temporal changes in different regions of underwater video frames. In contrast, our proposed method ensures consistency in enhancement within the same region and maintains spatial and temporal continuity between adjacent regions.
Conventional enhancement methods are not suitable for underwater video enhancement because of the significant variations in enhancement effects caused by the temporal changes in different regions over time, which severely affects the user experience. However, the Contrast Limited Adaptive Histogram Equalization (CLAHE)\cite{zuiderveld1994contrast} algorithm divides the image into $\mathbf{4\times 4 }$ small regions for individual enhancement. It employs bilinear interpolation in the central region, ensuring consistency in enhancement within the same region and maintaining spatial and temporal continuity between adjacent regions. Subsequently, we apply the Bayesian Retinex algorithm for fine-tuning the color.
For color correction, we employ the Bayesian Retinex algorithm, which first utilizes color correction methods to remove color casts and restore naturalness. Then, a multi-level gradient prior is established based on the color-corrected image. Color correction involves statistical methods to handle color shifts and can be calculated using the following formula:

\begin{equation}
U_{c}=\frac{255}{2}\times \left ( 1+\frac{S_{c}-M_{c}}{\mu \cdot V_{c}} \right )
\end{equation}

For an underwater RGB image, when  $\mathbf{c(R,G,B)}$, is computed for each of the three channels of the degraded underwater image  $\mathbf{S}$, $\mathbf{M_{c}}$ is the mean of the image  $\mathbf{S}$, and $\mathbf{V_{c}}$ is the variance of the image  $\mathbf{S}$. $\mathbf{\mu}$ is the parameter that regulates the underwater enhancement saturation, and for each color channel, we usually set it to 2.5
After the aforementioned computational processing, a constant calibration is applied to each color channel. Subsequently, the Bayesian Retinex\cite{fu2014retinex} image enhancement is performed. Using Bayesian inference, the Bayesian Retinex enhancement model simultaneously enhances the illumination component $\mathbf{L}$ and the reflectance component $\mathbf{R}$. The formulation of the posterior distribution in the Bayesian Retinex model can be expressed as:

\begin{equation}
p\left ( I,R\mid L  \right )\propto  p\left ( L\mid I,R \right ) p\left ( I \right ) p\left ( R \right ) 
\end{equation}

Meanwhile, the parameter $\mathbf{p}$ introduces a multi-order gradient prior to design $\mathbf{p(R)}$ in order to obtain a more complete underwater image structure. $\mathbf{p}$ in the first- and second-order gradients of the reflectivity, the image structure is richer, resulting in finer details. In pursuit of better spatial smoothness of illumination, a Gaussian distribution with zero mean and variance is used to model the first-order gradient prior of illumination, which is modeled as:

\begin{equation}
p_{1} =N\left ( \bigtriangledown I\mid 0,\sigma _{3}^{2} 1 \right ) 
\end{equation}

Also for the second-order gradient prior, for the approximation of the segmented linear component of the illumination, another Gaussian distribution with zero mean and variance $\mathbf{{{\sigma}}_{{4}}^{{2}}}$ is used to model the second-order derivative prior of the illumination:

\begin{equation}
p_{2} =N\left ( \bigtriangleup  I\mid 0,\sigma _{4}^{2} 1 \right ) 
\end{equation}

Ultimately, the priori $\mathbf{p(I)}$ is modeled as:

\begin{equation}
p\left ( I \right ) =p_{1} \left ( I \right ) p_{2} \left ( I \right )
\end{equation}

The final illumination adjustment is a fine-tuning operation performed on the environment, using an effective gamma correction to adjust the luminosity. The correction can be expressed as:

\begin{equation}
I_{e}=W\left ( \frac{I}{W}  \right )^{\frac{1}{\gamma } } 
\end{equation}

The channel $\mathbf{L_{e} }$ of the final image is calculated as:
\begin{equation}
L_{e} =I_{e} \cdot R
\end{equation}
\subsection{Keyframe Filtering Based on Pose and Image Quality}
Keyframe selection is a critical step in our approach, aiming to identify representative frames from a multitude of images while minimizing their impact on the subsequent 3-D reconstruction module. Diverging from conventional keyframe selection methods that rely solely on pose or image entropy criteria, our method combines both pose information and image sharpness to filter keyframes. This novel approach allows us to consider image quality alongside the appropriateness of keyframe placement.
\begin{equation}
\text{$P_c$} 
= \begin{bmatrix}
R & t \\
0 & 1
\end{bmatrix}
=\begin{bmatrix}
r_{1,1} & r_{1,2} & r_{1,3} & t_1 \\
r_{2,1} & r_{2,2} & r_{2,3} & t_2 \\
r_{3,1} & r_{3,2} & r_{3,3} & t_3 \\
0 & 0 & 0 & 1
\end{bmatrix}
\end{equation}
Where:
    $R$ is a 3$\times$3 rotation matrix representing the camera's orientation.
    $t$ is a 3$\times$1 translation vector representing the camera's position.
    The bottom-right element is always 1 to maintain correct matrix multiplication.\\
We calculate the angular difference and displacement between cameras using the following formulas:
\begin{equation}
\theta_{\Delta} = \arccos\left(1 - 0.5 \sqrt{(R_{13} - {R_{13}'}^2) + (R_{23} - {R_{23}'}^2) + (R_{33} - {R_{33}'}^2)}\right)
\end{equation}
\begin{equation}
Dis_{\Delta} = \sqrt{(R_{14} - {R_{14}'}^2) + (R_{24} - {R_{24}'}^2) + (R_{34} - {R_{34}'}^2)}
\end{equation}
Here, we use the camera-to-world (c2w) format for the pose matrices.\\
Additionally, we compute the image sharpness using the Laplacian operator to identify keyframes with higher quality in noisy datasets. We calculate the keyframe importance parameter as follows:
\begin{equation}
I = w_1 \times \text{sharpness} + w_2 \times Dis_{\Delta} + (1 - w_1 - w_2) \times \theta_{\Delta}
\end{equation}
We use a weighted average of sharpness and the angular difference with displacement to strike a balance between image quality and quantity. 
\subsection{3D Reconstruction Based on Instant-NGP}
\subsubsection{Neural Radiance Fields (NeRFs)}
Neural Radiance Fields (NeRFs) is a groundbreaking advancement in the field of 3D visual reconstruction. NeRFs was initially proposed by Ben Mildenhall et al\cite{mildenhall2021nerf}. in 2020 to address the challenging task of scene reconstruction and view synthesis from 2D images.
In essence, the original NeRF can be understood as a Multilayer Perceptron (MLP) that primarily consists of fully connected layers instead of convolutional layers. Its purpose is to learn a static 3D scene, often parameterized through the $\mathbf{MLPf_{\theta }:(x,d)\longrightarrow (c,\sigma )}$. The NeRF function takes as input a continuous representation of the scene, which is a $\mathbf{5D}$ vector containing a spatial 3D coordinate point x=(x,y,z) and the direction $\mathbf{d=(\theta, \phi)}$ from that coordinate position. The output of the function is the RGB color coordinates c = (r,g,b) of the 3D point and the corresponding opacity or density value $\mathbf{\sigma }$ at that location. The neural network can be represented as follows:

\begin{equation}
F_{\theta } :(x,d)\to (c,\sigma )
\end{equation}

The voxel density $\mathbf{\sigma \left ( x \right ) }$ can be understood as the probability that a ray traveling through space will be terminated by an infinitesimal particle at $\mathbf{x}$. This probability is differentiable and can be approximated as the opacity of the point at that location. Since the points on the observed ray of the camera along a particular direction are continuous, the color of the corresponding pixel in the imaging plane of that camera can be understood as the color integral of the point through which the corresponding ray passes can be expressed as:

\begin{equation}
C\left ( r \right ) =\int_{t_{n} }^{t_{f} } T(t)\cdot \sigma (r(t))\cdot c(r(t),d)dt
\end{equation}

By labeling the origin of a ray as $\mathbf{o}$ and the direction of the ray (i.e.the camera viewpoint) as $\mathbf{d}$, the ray can be represented as $\mathbf{r(t)=o+td}$, with the proximal and distal boundaries of $\mathbf{t}$ as $\mathbf{t_{n} }$ and $\mathbf{t_{f}}$. Respectively.
Where $\mathbf{T(t)}$ denotes the cumulative transparency of the section of the ray from $\mathbf{t_{n} }$ to $\mathbf{t}$, i.e.the probability that the ray has not been stopped by hitting any particle from $\mathbf{t_{n} }$ to $\mathbf{t}$, is denoted as:

\begin{equation}
T(t)=exp(-\int_{t_{n} }^{t} \sigma(r(s))ds
\end{equation}

In practical scenario applications, it is not possible to do the NeRF to estimate continuous 3D information, so a numerical approximation method, i.e.uniform random sampling method, is used, whose $\mathbf{i}$ sampling point can be expressed as:

\begin{equation}
t_{i}=U\left [ t_{n} +\frac{i-1}{N}(t_{f}-t_{n}, t_{n}+\frac{i}{N} ( t_{f}-t_{n} ) \right ]  
\end{equation}

The first step is to deal with the region on the ray that needs to be integrated by dividing the region into $\mathbf{N}$ parts, and numerical approximation of each small region ensures that the continuity of the adopted position, and simplifies the above color equation to:
\begin{equation}
\hat{C} (r)=\sum_{i=1}^{N} T_{i} \cdot (1-exp(-\sigma _{i} \cdot \delta _{_{i} } ))\cdot c_{i} 
\end{equation}

where the distance $\mathbf{T_{i} }$ between neighboring come sampling points can be expressed as:
\begin{equation}
T_{i} =exp(-\sum_{j=1}^{i-1}\sigma_{j}  \delta _{j}  ) 
\end{equation}

The rendering principle of NeRF is to sample and sum for each ray emitted by the camera. Hence NeRF's pain point: it is slightly less efficient. This is because arithmetic is still consumed for regions where rendering is not effective.
Ultimately the training loss of NeRF is directly determined with the $\mathbf{L2}$ loss of the rendering result, which can be expressed as:
\begin{equation}
L=\sum_{r\in R}^{}\left [ \parallel \hat{C_{c} } (r)-C(r) \parallel_{2}^{2}- \parallel\hat{C_{f} } (r)-C(r) \parallel_{2}^{2}\right ]  
\end{equation}

\subsubsection{Instant-NGP (Instant Neural Gradient Prediction)}
Instant-NGP\cite{tancik2023nerfstudio} is an innovative approach that addresses the challenge of gradient prediction in neural networks, particularly in the context of underwater scene reconstruction. It plays a fundamental role in enhancing the training stability and efficiency of our system.
\paragraph{Gradient Prediction Network (GPN)}
Instant-NGP introduces a Gradient Prediction Network (GPN), denoted as $\mathcal{N}_{\text{GPN}}$, which is a neural network responsible for predicting gradients. The GPN takes the network's current state $\mathbf{W}$ and an input sample $\mathbf{x}$ as input and predicts the gradient $\nabla \mathcal{L}(\mathbf{W}, \mathbf{x})$ with respect to the loss function $\mathcal{L}$. This can be expressed as:
\begin{equation}
\nabla \mathcal{L}(\mathbf{W}, \mathbf{x}) = \mathcal{N}_{\text{GPN}}(\mathbf{W}, \mathbf{x})
\end{equation}
\paragraph{Instantaneous Gradient Updates}
Once the gradient is predicted by the GPN, it is used to update the network's parameters $\mathbf{W}$ immediately. This enables the network to adapt rapidly to changing conditions, such as variations in underwater scenes. The instantaneous gradient update can be formulated as:
\begin{equation}
\mathbf{W}_{\text{new}} = \mathbf{W}_{\text{old}} - \alpha \cdot \nabla \mathcal{L}(\mathbf{W}_{\text{old}}, \mathbf{x})
\end{equation}
where $\alpha$ is the learning rate.\\
Our Underwater 3-D Reconstruction module is powered by the innovative Instant-NGP (Instant Neural Gradient Prediction)\cite{mueller2022instant} algorithm. This algorithm plays a pivotal role in accelerating the reconstruction process compared to traditional methods, such as Seathru-NeRF\cite{levy2023seathru}.
In the context of our underwater scene reconstruction system, the Underwater Reconstruction module harnesses the power of Instant-NGP to achieve both efficiency and accuracy. By incorporating this algorithm, we can reconstruct 3-D underwater scenes in a timely manner without compromising the quality of the reconstructions. This is paramount for tasks such as underwater navigation, environmental monitoring, and scientific exploration, where real-time or near-real-time feedback is essential.\\
In the following section (Section 5), we will present the experimental results and discuss the performance of our Underwater Reconstruction module in detail, showcasing how Instant-NGP contributes to the success of our system in accurately capturing the intricacies of underwater environments.

\section{Experiments and Results}
\subsection{Evaluation Metrics}
In this section, we outline the evaluation metrics used to assess the performance of our underwater scene reconstruction system, focusing on two key modules: the Underwater Enhancement module and the Underwater 3-D Reconstruction module.
\subsubsection{Underwater Enhancement Evaluation Metrics.}
We use three commonly used indicators for evaluating underwater images.
\paragraph{UICQE (Underwater Image Color Quality Enhancement)}\cite{yang2015underwater}
UICQE is a metric designed to evaluate the color enhancement quality of underwater images. It assesses the system's ability to improve the color fidelity and vibrancy of underwater scenes. Higher UICQE scores indicate superior color enhancement performance.
\paragraph{UIQM (Underwater Image Quality Measure)}\cite{panetta2015human}
UIQM measures the overall quality improvement achieved by the Underwater Enhancement module. It considers various aspects such as contrast, brightness, and color accuracy. A higher UIQM score signifies better overall image quality enhancement.
\paragraph{SSIM (Structural Similarity Index)}\cite{wang2004image}
SSIM is a widely used metric for assessing the structural similarity between the enhanced underwater images and their ground truth counterparts. It quantifies the preservation of image structures and details. A higher SSIM score indicates better preservation of structural information.
\subsubsection{Underwater 3-D Reconstruction Evaluation Metric.}
We use a similar metric as Levy et al\cite{levy2023seathru}. , which is widely used in the field of 3-D reconstruction.
\paragraph{PSNR (Peak Signal-to-Noise Ratio)}
 PSNR is a fundamental metric for evaluating the fidelity of 3-D reconstructed scenes compared to ground truth data. It quantifies the level of noise and distortion present in the reconstructed 3-D models. A higher PSNR value implies a closer match to the ground truth, indicating superior reconstruction accuracy.
\begin{equation}
\text{PSNR} = 10 \cdot \log_{10}\left(\frac{{\text{MAX}^2}}{{\text{MSE}}}\right)
\end{equation}
In this formula, PSNR stands for Peak Signal-to-Noise Ratio, MAX represents the maximum possible pixel value in the image (typically 255 for 8-bit grayscale images), and MSE denotes the Mean Squared Error, which measures the mean squared difference between the original image and the reconstructed image. PSNR is used to assess the quality of image reconstruction, with higher values indicating greater similarity between the reconstructed and original images.
\paragraph{Structural Similarity Index (SSIM)}\cite{wang2004image}
SSIM, as mentioned earlier, is also used in this context to evaluate the similarity between the reconstructed 3-D models and the ground truth. It assesses the preservation of structural details in the 3-D reconstructions.
\begin{equation}
\text{SSIM}(x, y) = \frac{{(2\mu_x\mu_y + C_1)(2\sigma_{xy} + C_2)}}{{(\mu_x^2 + \mu_y^2 + C_1)(\sigma_x^2 + \sigma_y^2 + C_2)}}
\end{equation}
In this formula:
    $x$ and $y$ are the two input images being compared.
    $\mu_x$ and $\mu_y$ represent the mean values of $x$ and $y$ respectively.
    $\sigma_x^2$ and $\sigma_y^2$ denote the variances of $x$ and $y$ respectively.
    $\sigma_{xy}$ represents the covariance between $x$ and $y$.
    $C_1$ and $C_2$ are small constants added to avoid division by zero. Typically, $C_1=(k_1L)^2$and $C_2=(k_2L)^2$ are used, where $L$ is the dynamic range of pixel values in the images (e.g., 255 for 8-bit images), and $k_1$ and $k_2$ are constants to control the impact of $C_1$ and $C_2$.

\subsection{Video Data}
The video data utilized in our study was obtained through underwater diving expeditions, capturing real-world underwater scenes. A total of 241 and 231 individual images were collected for our experiments, corresponding to two distinct underwater scenarios. These images were captured using GoPro HERO BLACK cameras, and were recorded at a resolution of 1280 x 720 pixels.
\subsection{Experimental Information}
Our experiments were conducted on a high-performance server equipped with a NVIDIA RTX 4090 GPU with 24GB of GPU memory and a Intel(R) Xeon(R) Platinum 8352V CPU @ 2.10GHz.After testing, in the 3D reconstruction module, we set the $aabb\_scale$ parameter to 32 to achieve the best training effect. Retain the default values for other parameters.
\subsection{Result and Discussion}

\subsubsection{Underwater Image Enhancement} 
Table 1 presents the results of our experiments, where we evaluated various methods using different evaluation metrics on our dataset.     We conducted experiments for five different cases (122, 167, 195, 214, and 216) to assess the performance of different methods in underwater 3D reconstruction.     For each case, we considered several methods: Fusion, UCM, RGHS, CLAHE, Seathru, and Our proposed method.     The evaluation metrics used include UCIQE, UIQM, and SSIM, which measure different aspects of image quality and reconstruction accuracy.     The metrics UCIQE and UIQM aim to quantify image quality, where higher values are generally better.     SSIM (Structural Similarity Index) measures the structural similarity between the reconstructed and ground truth images.     Based on the experimental results, it can be observed that our method performs well in some cases but does not necessarily excel in all metrics.    Higher SSIM values are desirable, and our method performs well in terms of SSIM, particularly achieving high SSIM values in some cases.     In some scenarios, other methods may achieve higher scores in terms of UCIQE and UIQM, indicating better image quality.  Please note that the final choice of method may depend on the specific circumstances and application requirements, as different methods may have strengths in different aspects. Overall, the experimental results suggest that our method is competitive in some scenarios but may not necessarily be the best choice in all situations.\\

\begin{figure}[htbp]
      \centering
      \includegraphics[scale=0.3]{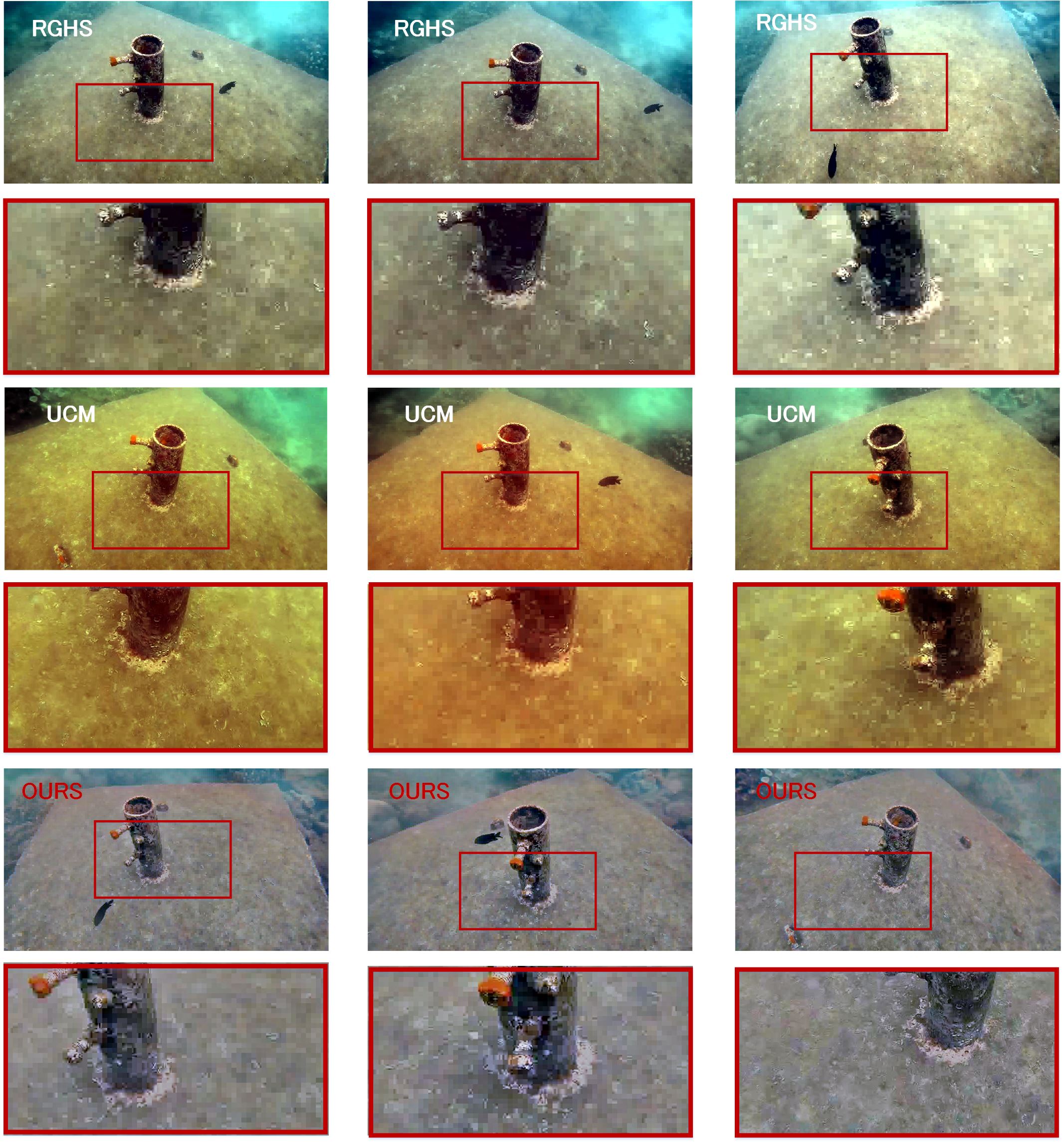}
      \caption{From top to bottom, each group is: RGHS, UCM, OURS.\\
(Since our method is to enhance the regions separately, our enhancement data is relatively consistent and not prone to sudden color changes. Compared to other methods, using overall enhancement, the image is prone to color mutations)}
\end{figure}

\begin{figure}[htbp]
      \centering
      \includegraphics[scale=0.11]{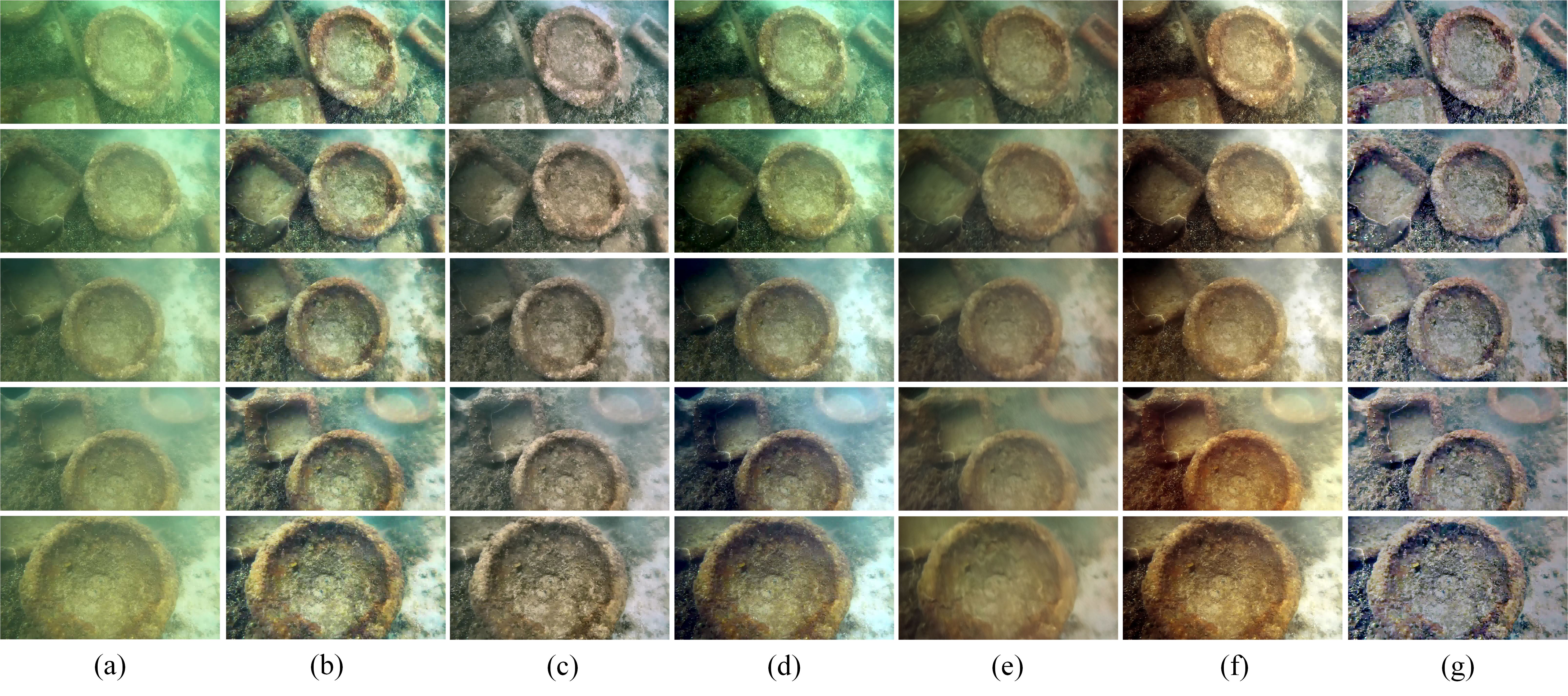}
      \caption{(a)Original (b)CLAHE (c)Fusion (d)RGHS (e)Seathru (f)UCM (g)Ours ("Seathru" is a proprietary algorithm, so we use the rendered images from "Seathru-NeRF" as its enhancement results in this context.)
}
\end{figure}

\begin{table}
\centering
\caption{Evaluation Metrics of Seathru-NeRF\cite{levy2023seathru} and our methods on our own dataset.\\(Bold the best data within each case, and italicize the second-best data.)}\label{tab1}
\begin{tabular}{p{2cm}p{3cm}p{2cm}p{2cm}p{2cm}}
\hline
Case & Methods & UCIQE &  UIQM & SSIM\\
\hline
\multirow{6}*{122}	&	Fusion	&	0.330	&	\emph{\textcolor{black}{0.569}}	&	\textbf{\textcolor{black}{0.872}}	\\
~	&	UCM	&	0.403	&	0.447	&	0.831	\\
~	&	RGHS	&	\textbf{\textcolor{black}{0.421}}	&	0.430	&	0.569	\\
~	&	CLAHE	&	0.358	&	0.514	&	\emph{\textcolor{black}{0.838}}	\\
~	&	Seathru	&	0.246	&	0.466	&	0.710	\\
~	&	Ours	&	\emph{\textcolor{black}{0.417}}	&	\textbf{\textcolor{black}{0.795}}	&	0.677	\\
\hline
\multirow{6}*{167}	&	Fusion	&	0.315	&	\emph{\textcolor{black}{0.546}}	&	\textbf{\textcolor{black}{0.909}}	\\
~	&	UCM	&	0.395	&	0.387	&	0.854	\\
~	&	RGHS	&	\emph{\textcolor{black}{0.410}}	&	0.298	&	\emph{\textcolor{black}{0.870}}	\\
~	&	CLAHE	&	0.350	&	0.467	&	0.859	\\
~	&	Seathru	&	0.273	&	0.471	&	0.753	\\
~	&	Ours	&	\textbf{\textcolor{black}{0.412}}	&	\textbf{\textcolor{black}{0.709}}	&	0.691	\\
\hline
\multirow{6}*{195}	&	Fusion	&	0.290	&	\emph{\textcolor{black}{0.534}}	&	\textbf{\textcolor{black}{0.925}}	\\
~	&	UCM	&	\textbf{\textcolor{black}{0.399}}	&	0.377	&	0.894	\\
~	&	RGHS	&	\emph{\textcolor{black}{0.398}}&	0.409	&	\emph{\textcolor{black}{0.901}}	\\
~	&	CLAHE	&	0.329	&	0.459	&	0.876	\\
~	&	Seathru	&	0.279	&	0.421	&	0.803	\\
~	&	Ours	&	0.378	&	\textbf{\textcolor{black}{0.647}}	&	0.725	\\
\hline
\multirow{6}*{214}	&	Fusion	&	0.298	&	\emph{\textcolor{black}{0.523}}	&	\textbf{\textcolor{black}{0.907}}	\\
~	&	UCM	&	\emph{\textcolor{black}{0.416}}	&	0.276	&	0.860	\\
~	&	RGHS	&	\textbf{\textcolor{black}{0.435}}	&	0.496	&	0.843	\\
~	&	CLAHE	&	0.324	&	0.448	&	\emph{\textcolor{black}{0.872}}	\\
~	&	Seathru	&	0.264	&	0.411	&	0.795	\\
~	&	Ours	&	0.389	&	\textbf{\textcolor{black}{0.620}}	&	0.750	\\
\hline
\multirow{6}*{216}	&	Fusion	&	0.311	&	0.507	&	\emph{\textcolor{black}{0.879}}	\\
~	&	UCM	&	\emph{\textcolor{black}{0.412}}	&	0.368	&	0.873	\\
~	&	RGHS	&	\textbf{\textcolor{black}{0.427}}	&	0.489	&	\textbf{\textcolor{black}{0.900}}	\\
~	&	CLAHE	&	0.343	&	\emph{\textcolor{black}{0.537}}	&	0.847	\\
~	&	Seathru	&	0.286	&	0.178	&	0.747	\\
~	&	Ours	&	0.406	&	\textbf{\textcolor{black}{0.714}}	&	0.702	\\
\hline
\end{tabular}
\end{table}

\subsubsection{Underwater Reconstruction} 
Table 2 presents the results of our experiments, comparing Seathru-NeRF with our proposed methods using different evaluation metrics on our own dataset. We evaluated two versions of our method, one with a runtime of approximately 1 minute and another with a runtime of approximately 10 minutes. In terms of PSNR (Peak Signal-to-Noise Ratio), the first version of our method achieved the highest score of 18.43, highlighted in bold and red, indicating superior performance in terms of image fidelity. For SSIM (Structural Similarity Index), Seathru-NeRF with a runtime of approximately 7 hours obtained the highest score of 0.6958, also highlighted in bold and red, suggesting better structural similarity with ground truth images. It's worth noting that the trade-off between runtime and performance is evident, as our faster version achieved a competitive PSNR score, while Seathru-NeRF with significantly longer runtime excelled in SSIM. These results indicate that our methods offer a good balance between runtime efficiency and reconstruction quality, making them suitable for practical applications.  However, the choice between the two methods may depend on specific use cases and priorities. Overall, our methods demonstrate promising results in terms of both PSNR and SSIM, showcasing their effectiveness in underwater scene reconstruction.
\begin{table}
\centering
\caption{Evaluation Metrics of Seathru-NeRF and our methods on our own dataset.\\(Bold the best data within each case)}\label{tab2}
\begin{tabular}{p{5cm}p{3cm}p{3cm}}
\hline
Algorithm &  PSNR & SSIM\\
\hline
Ours($ \sim $ 1min) &  \textbf{\textcolor{black}{18.43}} & 0.6610\\
Ours($ \sim $ 10min) &  18.40 & 0.6677\\
Seathru-NeRF($ \sim $ 7hours) & 16.54 & \textbf{\textcolor{black}{0.6958}}\\
Seathru-NeRF($ \sim $ 20hours) & 14.97 & 0.6676\\
\hline
\end{tabular}
\end{table}
\begin{figure}[htbp]
      \centering
      \includegraphics[scale=0.2]{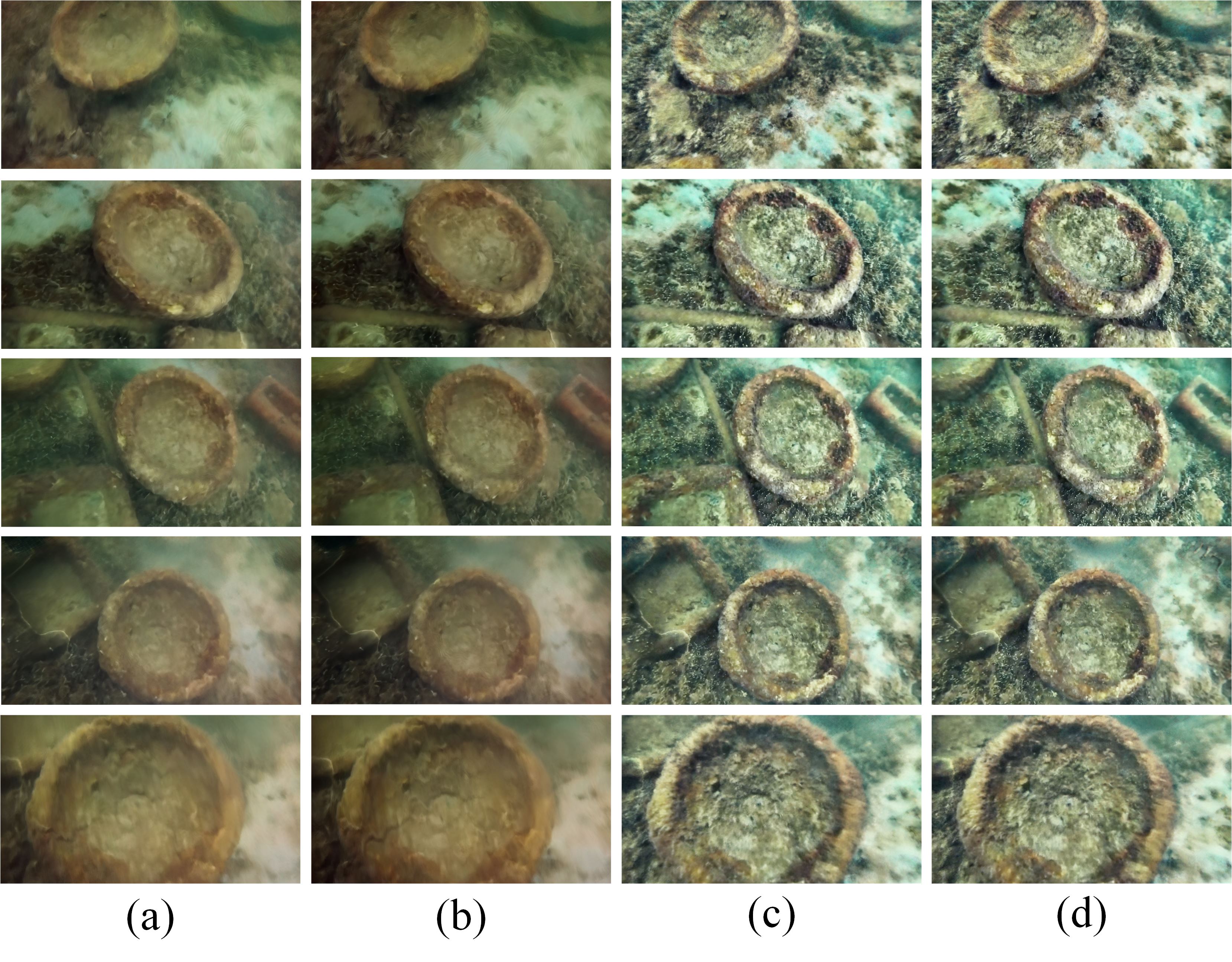}
      \caption{(a)Seathru-NeRF$ \sim $ 7h (b)Seathru-NeRF$ \sim $ 20h (c)Ours$ \sim $ 1min (d)Ours$ \sim $ 10min
}
\end{figure}
\subsection{Limitations}
While our method achieves good reconstruction results with low computational time, it is important to note that, as an image enhancement rather than image restoration approach, there is still room for improvement in terms of color fidelity. Additionally, during our experiments, we observed that the omission of modeling underwater imaging effects still leads to the formation of artifacts during training. Therefore, we consider optimizing the existing method to enhance reconstruction color accuracy and quality as a future avenue of research.

\section{Conclusion And Future Work}
We provide an extension to 3D visual reconstruction for challenging scenes that enables NeRF development in underwater scenes and scattering media. Our work provides a complete underwater 3D reconstruction system to address this challenge, and our approach is able to discard the drawbacks of time-consuming training and inefficient rendering of the original NeRF. Our system can be utilized to achieve fast, high-performance underwater 3D reconstruction.\\
Our image enhancement method relies on the chunking idea of CLAHE and Bayesian Retinux, which avoids a series of problems caused by sudden color changes, and our method makes the continuity of image data good. We utilize Colmap visual tool to get the bitmap information of the original image frames, which has higher robustness. At the same time, our key frame filtering module is highly efficient, which greatly saves arithmetic resources and improves the efficiency of the whole system synchronously. Our chosen Instant-ngp method utilizes hash coding to make NeRF 3D reconstruction much more efficient. As a result, the overall performance of our system is better.\\
However, there are several shortcomings of our method, although the overall performance of our system is superior, our system requires prior extraction of the image's bitmap information as input, which may be challenging in more challenging scenarios. Our method achieves better reconstruction results and is less time consuming. However, there is still room for improvement in color fidelity since we are using an image enhancement method, not an image restoration method. In our experiments, we found that many artifacts are still formed during the training process due to the neglect of underwater imaging modeling. In the future, we will continue to optimize our system, optimize the existing methods to improve the reconstruction color accuracy and reconstruction quality, and continue to explore the 3D reconstruction of challenging scenes.This work also have potential values in some other fields as robotics.\cite{liu2019lifelong,liu2019federated,liu2021peer,zheng2021applications,zhang2022authros,liu2017singular,10347007}

\subsubsection{Acknowledgements} This work was supported by National Natural Science Foundation of China (NSFC) (Grant No. 62162024, 62162022), the Major science and technology project of Hainan Province (Grant No. ZDKJ2020012), Hainan Provincial Natural Science Foundation of China (Grant No. 620MS021), Youth Foundation Project of Hainan Natural Science Foundation (621QN211)

%
%
%
\bibliographystyle{splncs04}
\bibliography{mybibliography}

\begin{thebibliography}{10}
\providecommand{\url}[1]{\texttt{#1}}
\providecommand{\urlprefix}{URL }
\providecommand{\doi}[1]{https://doi.org/#1}

\bibitem{dube2017segmatch}
Dub{\'e}, R., Dugas, D., Stumm, E., Nieto, J., Siegwart, R., Cadena, C.: Segmatch: Segment based place recognition in 3d point clouds. In: 2017 IEEE International Conference on Robotics and Automation (ICRA). pp. 5266--5272. IEEE (2017)

\bibitem{7025927}
Fu, X., Zhuang, P., Huang, Y., Liao, Y., Zhang, X.P., Ding, X.: A retinex-based enhancing approach for single underwater image. In: 2014 IEEE International Conference on Image Processing (ICIP). pp. 4572--4576 (2014). \doi{10.1109/ICIP.2014.7025927}

\bibitem{fu2014retinex}
Fu, X., Zhuang, P., Huang, Y., Liao, Y., Zhang, X.P., Ding, X.: A retinex-based enhancing approach for single underwater image. In: 2014 IEEE international conference on image processing (ICIP). pp. 4572--4576. IEEE (2014)

\bibitem{guo2022nerfren}
Guo, Y.C., Kang, D., Bao, L., He, Y., Zhang, S.H.: Nerfren: Neural radiance fields with reflections. In: Proceedings of the IEEE/CVF Conference on Computer Vision and Pattern Recognition. pp. 18409--18418 (2022)

\bibitem{6522017}
Hitam, M.S., Awalludin, E.A., Jawahir Hj Wan~Yussof, W.N., Bachok, Z.: Mixture contrast limited adaptive histogram equalization for underwater image enhancement. In: 2013 International Conference on Computer Applications Technology (ICCAT). pp.~1--5 (2013). \doi{10.1109/ICCAT.2013.6522017}

\bibitem{klein2007parallel}
Klein, G., Murray, D.: Parallel tracking and mapping for small ar workspaces. In: 2007 6th IEEE and ACM international symposium on mixed and augmented reality. pp. 225--234. IEEE (2007)

\bibitem{konolige2008frameslam}
Konolige, K., Agrawal, M.: Frameslam: From bundle adjustment to real-time visual mapping. IEEE Transactions on Robotics  \textbf{24}(5),  1066--1077 (2008)

\bibitem{levy2023seathru}
Levy, D., Peleg, A., Pearl, N., Rosenbaum, D., Akkaynak, D., Korman, S., Treibitz, T.: Seathru-nerf: Neural radiance fields in scattering media. In: Proceedings of the IEEE/CVF Conference on Computer Vision and Pattern Recognition. pp. 56--65 (2023)

\bibitem{liu2017singular}
Liu, B., Cheng, J., Cai, K., Shi, P., Tang, X.: Singular point probability improve lstm network performance for long-term traffic flow prediction. In: Theoretical Computer Science: 35th National Conference, NCTCS 2017, Wuhan, China, October 14-15, 2017, Proceedings. pp. 328--340. Springer (2017)

\bibitem{liu2021peer}
Liu, B., Wang, L., Chen, X., Huang, L., Han, D., Xu, C.Z.: Peer-assisted robotic learning: a data-driven collaborative learning approach for cloud robotic systems. In: 2021 IEEE International Conference on Robotics and Automation (ICRA). pp. 4062--4070. IEEE (2021)

\bibitem{liu2019lifelong}
Liu, B., Wang, L., Liu, M.: Lifelong federated reinforcement learning: a learning architecture for navigation in cloud robotic systems. IEEE Robotics and Automation Letters  \textbf{4}(4),  4555--4562 (2019)

\bibitem{10347007}
Liu, B., Wang, L., Liu, M.: Roboec2: A novel cloud robotic system with dynamic network offloading assisted by amazon ec2. IEEE Transactions on Automation Science and Engineering pp. 1--15 (2023). \doi{10.1109/TASE.2023.3305522}

\bibitem{liu2019federated}
Liu, B., Wang, L., Liu, M., Xu, C.Z.: Federated imitation learning: A novel framework for cloud robotic systems with heterogeneous sensor data. IEEE Robotics and Automation Letters  \textbf{5}(2),  3509--3516 (2019)

\bibitem{martinbrualla2020nerfw}
Martin-Brualla, R., Radwan, N., Sajjadi, M.S.M., Barron, J.T., Dosovitskiy, A., Duckworth, D.: {NeRF in the Wild: Neural Radiance Fields for Unconstrained Photo Collections}. In: CVPR (2021)

\bibitem{mildenhall2021nerf}
Mildenhall, B., Srinivasan, P.P., Tancik, M., Barron, J.T., Ramamoorthi, R., Ng, R.: Nerf: Representing scenes as neural radiance fields for view synthesis. Communications of the ACM  \textbf{65}(1),  99--106 (2021)

\bibitem{mueller2022instant}
M\"uller, T., Evans, A., Schied, C., Keller, A.: Instant neural graphics primitives with a multiresolution hash encoding. ACM Trans. Graph.  \textbf{41}(4),  102:1--102:15 (Jul 2022). \doi{10.1145/3528223.3530127}, \url{https://doi.org/10.1145/3528223.3530127}

\bibitem{mur2015orb}
Mur-Artal, R., Montiel, J.M.M., Tardos, J.D.: Orb-slam: a versatile and accurate monocular slam system. IEEE transactions on robotics  \textbf{31}(5),  1147--1163 (2015)

\bibitem{panetta2015human}
Panetta, K., Gao, C., Agaian, S.: Human-visual-system-inspired underwater image quality measures. IEEE Journal of Oceanic Engineering  \textbf{41}(3),  541--551 (2015)

\bibitem{pumarola2021d}
Pumarola, A., Corona, E., Pons-Moll, G., Moreno-Noguer, F.: D-nerf: Neural radiance fields for dynamic scenes. In: Proceedings of the IEEE/CVF Conference on Computer Vision and Pattern Recognition. pp. 10318--10327 (2021)

\bibitem{revaud2015epicflow}
Revaud, J., Weinzaepfel, P., Harchaoui, Z., Schmid, C.: Epicflow: Edge-preserving interpolation of correspondences for optical flow. In: Proceedings of the IEEE conference on computer vision and pattern recognition. pp. 1164--1172 (2015)

\bibitem{schoenberger2016sfm}
Sch\"{o}nberger, J.L., Frahm, J.M.: Structure-from-motion revisited. In: Conference on Computer Vision and Pattern Recognition (CVPR) (2016)

\bibitem{schoenberger2016mvs}
Sch\"{o}nberger, J.L., Zheng, E., Pollefeys, M., Frahm, J.M.: Pixelwise view selection for unstructured multi-view stereo. In: European Conference on Computer Vision (ECCV) (2016)

\bibitem{tancik2023nerfstudio}
Tancik, M., Weber, E., Ng, E., Li, R., Yi, B., Wang, T., Kristoffersen, A., Austin, J., Salahi, K., Ahuja, A., et~al.: Nerfstudio: A modular framework for neural radiance field development. In: ACM SIGGRAPH 2023 Conference Proceedings. pp. 1--12 (2023)

\bibitem{6241430}
Treibitz, T., Schechner, Y.Y.: Turbid scene enhancement using multi-directional illumination fusion. IEEE Transactions on Image Processing  \textbf{21}(11),  4662--4667 (2012). \doi{10.1109/TIP.2012.2208978}

\bibitem{wang2004image}
Wang, Z., Bovik, A.C., Sheikh, H.R., Simoncelli, E.P.: Image quality assessment: from error visibility to structural similarity. IEEE transactions on image processing  \textbf{13}(4),  600--612 (2004)

\bibitem{yang2015underwater}
Yang, M., Sowmya, A.: An underwater color image quality evaluation metric. IEEE Transactions on Image Processing  \textbf{24}(12),  6062--6071 (2015)

\bibitem{zhang2020nerf++}
Zhang, K., Riegler, G., Snavely, N., Koltun, V.: Nerf++: Analyzing and improving neural radiance fields. arXiv preprint arXiv:2010.07492  (2020)

\bibitem{zhang2022authros}
Zhang, S., Li, W., Li, X., Liu, B.: Authros: Secure data sharing among robot operating systems based on ethereum. In: 2022 IEEE 22nd International Conference on Software Quality, Reliability and Security (QRS). pp. 147--156. IEEE (2022)

\bibitem{zheng2021applications}
Zheng, Z., Zhou, Y., Sun, Y., Wang, Z., Liu, B., Li, K.: Applications of federated learning in smart cities: recent advances, taxonomy, and open challenges. Connection Science pp. 1--28 (2021)

\bibitem{zuiderveld1994contrast}
Zuiderveld, K.: Contrast limited adaptive histogram equalization. Graphics gems pp. 474--485 (1994)

\end{thebibliography}

\end{document}